\documentclass[conference]{IEEEtran}
\IEEEoverridecommandlockouts
\usepackage{cite}
\usepackage{amsmath,amssymb,amsfonts}
\usepackage{graphicx}
\usepackage{textcomp}
\usepackage{xcolor}
\usepackage{url}
\usepackage{booktabs}
\usepackage{algorithm}
\usepackage{multirow}
\usepackage{algpseudocode}

\def\BibTeX{{\rm B\kern-.05em{\sc i\kern-.025em b}\kern-.08em
    T\kern-.1667em\lower.7ex\hbox{E}\kern-.125emX}}

\newcommand{\mymethod}{MIAR}
\newcommand{\RED}[1]{\textcolor{red}{#1}}  % Bold and red To-Do command
\newcommand{\BLUE}[1]{\textcolor{blue}{#1}}  % Bold and red To-Do command

\begin{document}

\title{MIAR: Medical Image Super-Resolution With Autoregressive Modeling
\vspace{-0.5em}}

\author{\IEEEauthorblockN{Fang Li$^{1}$,
Yinglong Li$^{1}$,
Hongyu Wu$^{1}$\IEEEauthorrefmark{2}, 
Yang Gao$^{1}$, 
Minwei Zhao$^{2}$,
Aimin Hao$^{1}$}
\IEEEauthorblockA{$^{1}$State Key Laboratory of Virtual Reality Technology and Systems, QRI $\&$ SCSE, Beihang University}
\IEEEauthorblockA{$^{2}$Department of Orthopaedics, Peking University Third Hospital}

% <-this % stops an unwanted space
\thanks{ \IEEEauthorrefmark{2}Corresponding author.

}
\vspace{-0.5em}
}

% \author{Anonymous ICME submission}

\maketitle

\begin{abstract}
Medical Image Super-Resolution (MISR) aims to enhance spatial resolution without necessitating hardware modifications. Although deep learning has yielded promising results, existing paradigms face a critical trade-off: diffusion-based methods suffer from prohibitive inference latency and compromised structural fidelity, whereas regression-based models typically yield over-smoothed results lacking perceptual realism. To address these limitations, we propose \mymethod, which reformulates super-resolution as a conditional, progressive next-scale prediction task via a multi-scale autoregressive framework. To ensure structural fidelity, we augment the autoregressive backbone with a Scale-Adaptive Structural Decoder. Furthermore, we integrate a hierarchical beam search strategy during inference to mitigate the recursive error accumulation inherent in autoregressive generation—a phenomenon especially pronounced  in medical images. Extensive experiments demonstrate that \mymethod{} establishes new state-of-the-art benchmarks while maintaining superior fidelity. Notably, our framework achieves a 7.86\% improvement in the perceptual metric MUSIQ relative to the state-of-the-art, while simultaneously delivering a 2.02$\times$ speedup over diffusion-based methods. Code is available at \url{https://github.com/Neesky/MIAR}
\end{abstract}

\section{Introduction}

Medical imaging modalities, such as magnetic resonance imaging (MRI) and cone-beam computed tomography (CBCT), are indispensable tools in clinical diagnosis, disease staging, and preoperative planning. These techniques enable non-invasive acquisition of high-resolution(HR) anatomical and functional information. However, physical and physiological constraints often limit the achievable spatial resolution. Specifically, MRI requires prolonged acquisition times, which not only reduce clinical throughput but also increase patient discomfort and susceptibility to motion-induced artifacts~\cite{feng2022multimodal}. On the other hand, CBCT typically operates under low-dose protocols and hardware limitations, resulting in intrinsically low-resolution (LR) volumes.

To address these limitations, super-resolution reconstruction has emerged as a promising computational technique. Conventional SR methods struggle to accurately model the complex nonlinear relationships between low-resolution and high-resolution images, often failing to recover high-frequency anatomical details critical for clinical interpretation~\cite{khaledyan2020low}. 

Recent advances in deep learning have led to the development of SR methods, particularly those based on convolutional neural networks (CNNs)~\cite{zhang2021mr} and Transformer-based architectures~\cite{ji2024deform}, which significantly improve reconstruction performance. Nonetheless, these models may still produce over-smoothed outputs and struggle to recover fine anatomical structures, especially in images with complex textures or pathological variations. Generative models, including generative adversarial networks (GANs)~\cite{li2021high}, have been employed to improve perceptual realism, but they often suffer from training instability and mode collapse.

Diffusion models (DMs) have recently emerged as a compelling alternative for medical image SR~\cite{li2024rethinking,mao2023disc}, owing to their ability to model complex data distributions via iterative denoising from Gaussian noise. Compared to GANs, DMs offer improved generation stability and fidelity. However, their inference process typically requires hundreds of iterative steps, which imposes high computational costs and limits clinical applicability. Moreover, reducing the number of inference steps to accelerate generation may introduce artificial structures or distortions that deviate from true anatomical representations, thus compromising diagnostic reliability.

\begin{figure*}[t]
\vspace{-10px}
\centering
\includegraphics[width=0.95\linewidth]{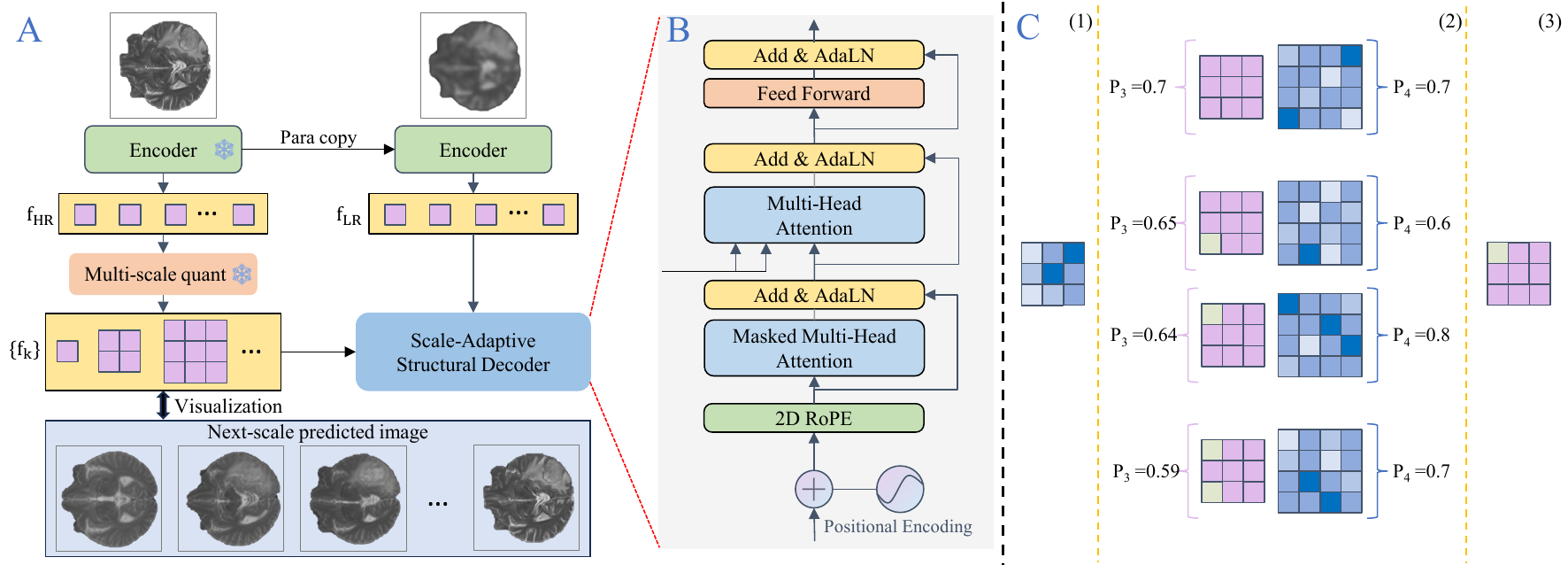}
\caption{\textbf{ Overview of the \mymethod{}.} 
(A) MSVQ-VAE: a HR image is decomposed into multi-scale discrete latent codes $\{f_k\}$ via a MSVQ-VAE, while the LR input is simultaneously encoded into continuous representations $f_{LR}$ using a pre-trained encoder. The bottom panel visualizes the progressive, coarse-to-fine anatomical reconstruction across successive scales.
(B) Scale-Adaptive Structural Decoder: This module integrates $\{f_k\}$ and $f_{LR}$ into an optimized decoder-style autoregressive generator, facilitating spatially-aware and scale-consistent synthesis.
(C) Hierarchical Beam Search: A hierarchical beam search strategy with $2^n$ candidate paths is introduced to mitigate error accumulation during autoregressive token generation across multiple spatial layers.}
\label{fig:framework}
\vspace{-10px}
\end{figure*}

Autoregressive models have emerged as a scalable alternative to diffusion models in generative tasks, offering a more efficient framework for image synthesis. Notable implementations\cite{sun2024autoregressive}, which leverage discrete token prediction for image generation. This paradigm enables autoregressive models to produce high-quality images while avoiding the iterative process inherent in diffusion models, thus reducing computational complexity. Recently, VAR \cite{tian2024visual} has attracted considerable attention by innovating the quantization of images into scale-wise token maps and generating images through next-scale prediction, achieving impressive results across a wide range of generative tasks. This novel architecture ensures fine-grained detail and fidelity by employing a progressive generation strategy. Additionally, VAR reduces the number of inference steps compared to traditional diffusion models, thereby enhancing computational efficiency.

However, Image Super-Resolution in medical imaging poses several unique challenges: 
1) How to effectively condition on LR inputs to reconstruct anatomically accurate and high-resolution outputs; 
2) In autoregressive models that linearize image data into one-dimensional token sequences, how to better encode spatial dependencies and positional relationships among visual tokens; 
3) Due to the high structural similarity across medical images, autoregressive prediction is prone to cumulative error propagation, which can degrade reconstruction quality over successive steps.

To address these challenges, we propose a novel Next-Scale Prediction Autoregressive Modeling framework for Medical Image Super-Resolution (MISR). Built upon the Visual Autoregressive (VAR) architecture, our method introduces  Scale-Adaptive Structural Decoder to more effectively capture semantic information from low-resolution inputs while preserving spatial coherence across the token sequence. Additionally, to alleviate error accumulation during autoregressive inference, we introduce a hierarchical beam search strategy that expands the hypothesis space at each prediction step, leading to more stable and anatomically consistent reconstructions.

Our main contributions are summarized as follows:

\begin{itemize}
     \item We present \mymethod{}, a novel autoregressive framework that integrates a Scale-Adaptive Structural Decoder into the VAR pipeline. By encoding LR priors and leveraging 2D Rotary Positional Embeddings (RoPE), our framework significantly enhances spatial coherence and preserves structural fidelity during generation.

    \item To mitigate error accumulation during sequential inference, we incorporate a hierarchical beam search sampling strategy that explores multiple high-probability generation paths. This improves robustness against hierarchical prediction errors.

    \item Through both quantitative and qualitative analyses, we demonstrate that \mymethod{} exhibits strong performance, achieving high-quality and realistic image generation while also showcasing the promising generation speed.
\end{itemize}

\section{Methodology}

\mymethod{} tackles the inherent challenges of autoregressive medical image super-resolution via three synergistic components: (i) a multi-scale quantization framework, (ii) a Scale-Adaptive Structural Decoder for precise conditioning, and (iii) a hierarchical search strategy to facilitate robust inference and high-fidelity inference.

\subsection{Multi-Scale Vector Quantized Variational Autoencoder}
To facilitate autoregressive modeling across spatial scales, we decompose the HR image, $\text{img}_{HR} \in \mathbb{R}^{H \times W \times C}$, into a hierarchy of discrete latent tokens using a  Multi-Scale Vector Quantized Variational Autoencoder (MSVQ-VAE). Unlike conventional single-scale quantization, this paradigm represents the image as a sequence of grids $(r_1, r_2, \dots, r_S)$, where $r_1$ is a $1 \times 1$ base token grid. Specifically, the image is first mapped into a continuous latent space $f_{HR} \in \mathbb{R}^{C_{HR} \times L_{HR} \times L_{HR}}$ via an encoder. The multi-scale residual quantization process is then recursively defined as:

\begin{equation}
f_{k}=f_{HR}-\sum_{m=1}^{k-1} \text{upsample}\left(\text{lookup}\left(Z, r_{m}\right)\right)
\end{equation}

where $f_k$ denotes the multi-scale feature map at scale $k$, which is progressively refined by the residual contributions of preceding scales. For any spatial position $(i, j)$ at a given scale, the discrete token mapping $q^{(i, j)}$ is determined via a nearest-neighbor search within the codebook $Z$:

\begin{equation}
q^{(i,j)} = \left( \arg \min_{v \in [V]} \left\| \text{lookup}(Z, v) - f^{(i,j)} \right\|_2 \right) \in [V]
\end{equation}

Here, $V$ denotes the codebook size. This hierarchical structure allows the model to capture global semantic structures at coarser scales while progressively recovering fine-grained anatomical details. To mitigate the risk of the model over-relying on deep architectural layers and to strengthen early-stage conditioning, we adopt a scale-aware dropout mechanism during training~\cite{li2024imagefolder}. By stochastically omitting higher scale in the generation path, the model is constrained to leverage lower-scale information and the LR prior $f_{LR}$ for subsequent predictions. This strategy compels the entire decoding hierarchy to participate meaningfully in the reconstruction task from the onset.

\subsection{Scale-Adaptive Structural Decoder}

Integrating structural priors into autoregressive (AR) frameworks remains a non-trivial challenge, particularly in medical image super-resolution where anatomical precision is paramount. While diffusion-based models leverage architectural priors like ControlNet, such mechanisms are often incompatible with the sequential nature of AR transformers. Existing prefix-based conditioning strategies, such as VARSR~\cite{qu2025visual}, frequently suffer from accumulated sampling errors and semantic drift, leading to incoherent token generation and degraded reconstruction quality.

To address these limitations, we propose the Scale-Adaptive Structural Decoder, which reformulates  HR synthesis as a conditional multi-scale autoregressive factorization:
\begin{equation}
p(r_1, r_2, \dots, r_K \mid f_{LR}) = \prod_{k=1}^K p(r_k \mid r_1, \dots, r_{k-1}, f_{LR})
\end{equation}

The architectural implementation consists of two primary stages. First, the LR input $\text{img}_{LR} \in \mathbb{R}^{H \times W \times C}$ is mapped to a continuous latent space $f_{LR} \in \mathbb{R}^{C_{LR} \times L_{LR} \times L_{LR}}$ via an encoder initialized with pretrained weights. This encoder is initialized with pretrained weights to ensure representational consistency between the LR and HR latent spaces. 

Second, the structural prior $f_{LR}$ is integrated into the generative process within the Scale-Adaptive Structural Decoder (Fig.~\ref{fig:framework}B). We utilize the decoder where $f_{LR}$ serves as the keys and values for cross-attention layers. while the multi-scale feature map before  the current scale, $\{f_k\}$, acts as the query. This enables the model to dynamically interrogate the structural prior at each scale, ensuring that the generated HR tokens remain spatially and semantically grounded. To further preserve anatomical fidelity, 2D RoPE are incorporated throughout the decoding pipeline to maintain precise spatial correspondences across the hierarchy.

\subsection{Hierarchical Beam Search}

\begin{figure*}[t] 
\centering 
\includegraphics[width=0.9\linewidth]{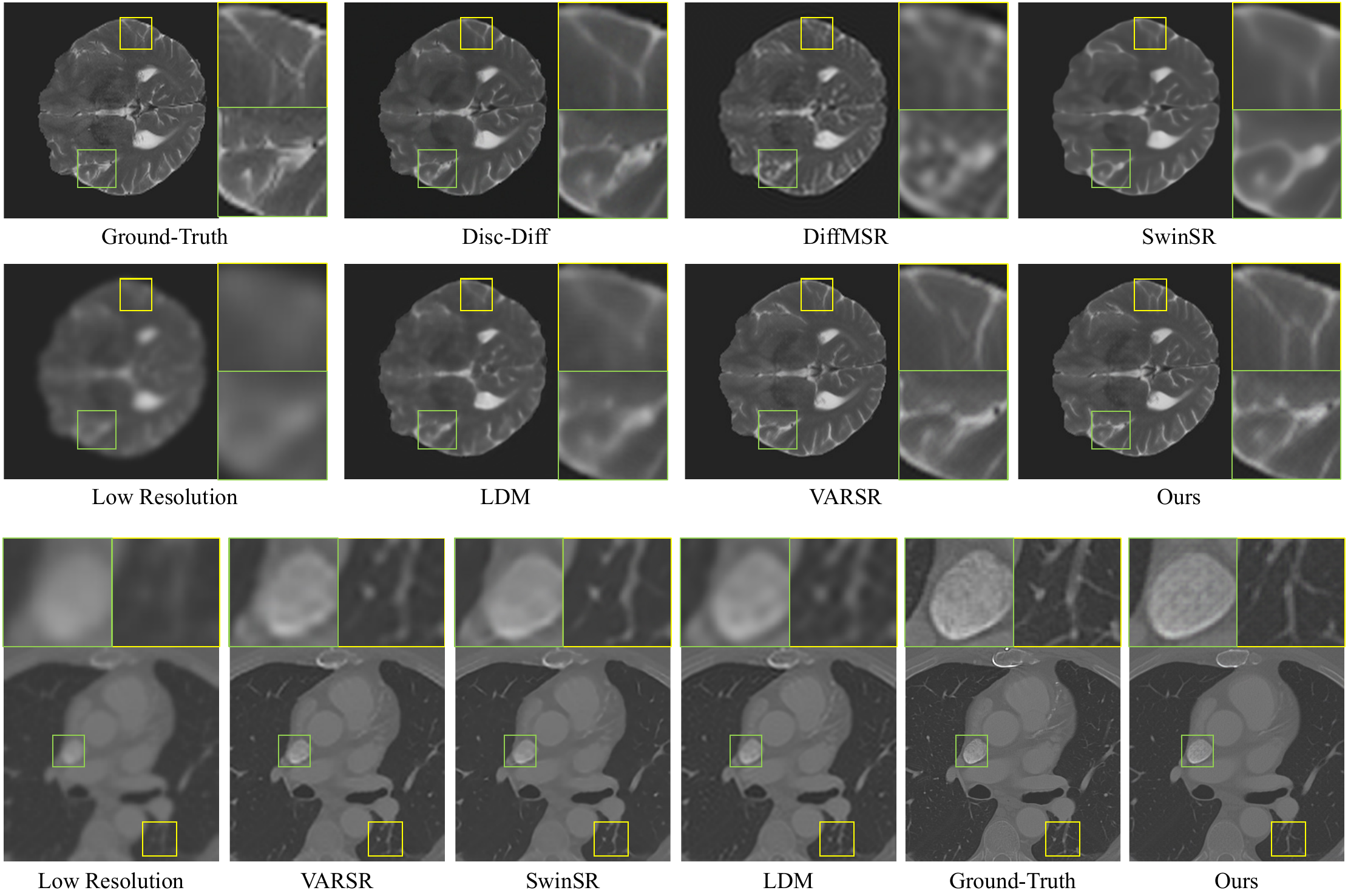}
\caption{\textbf{Visual Experiment.} We conduct a comprehensive evaluation of our proposed method, \mymethod{}, against state-of-the-art super-resolution approaches on two medical imaging datasets: brain MRI and lung CT. For the brain dataset, we include both multimodal (Disc-Diff~\cite{mao2023disc}, DiffMSR~\cite{li2024rethinking}) and unimodal (SwinIR~\cite{liang2021swinir}, LDM~\cite{rombach2022high}, VARSR~\cite{qu2025visual}) baselines, while for the lung dataset, we focus on unimodal settings due to the modality constraints. Our approach consistently achieves superior perceptual fidelity as verified through qualitative visual assessment.
}
\vspace{-10px}
\label{fig:contrast}
\end{figure*}

Due to the structural similarities and local continuity of anatomical features in medical images, the predicted tokens in medical image synthesis often exhibit high numerical proximity (Sec. \ref{subsec:beam_search}). This increases the risk of error accumulation in autoregressive generation, especially when employing widely-used sampling strategies such as top-$k$ and top-$p$. Hierarchical Beam search is a well-established decoding algorithm that mitigates this issue by retaining multiple candidate sequences during inference. It can be viewed as a pruned version of the classic breadth-first search, constrained by a fixed beam width. While standard autoregressive models for text generation predict a single index at each step, next-scale prediction autoregressive models must generate a grid of \( h_i \times h_i \) tokens at each scale. To address this problem, we propose hierarchical beam search specifically designed for next-scale prediction in medical image super-resolution.

As shown in Alg. \ref{alg:beam}, the process is as follows: First, the candidate set \( B_0 \), initialized with the start symbol, is created. At each generation step $i$, the model selects $n$ candidate tokens based on a predefined beam width. These candidates are used to construct $2^n$ possible token configurations, which are evaluated using a scoring function. The top-scoring configuration is propagated to the next step, ensuring both optimality and diversity in the autoregressive decoding process.

\begin{algorithm}
\caption{Hierarchical Beam Search in VAR}
\begin{algorithmic}[1]
\State \textbf{Input:} beam width $n$, max steps $T$
\State Initialize beam: $B_1 \gets \{ \langle 0, \texttt{ST} \rangle \}$
\For{$i = 1$ to $T$}
    \State $B_{i+1} \gets \emptyset$
    \State $\mathcal{V}_{choose}(i) \gets \texttt{chooseTokens}(B_i, n)$
    \State $Y \gets \texttt{getCandidate}(\mathcal{V}_{choose}(i))$
    \For{$y_i \in Y$} \Comment{$2^n$ choices}
        \State $s' \gets \texttt{score}(s, y_i)$
        \State $B_{i+1} \gets B_{i+1} \cup \langle s', {y_{1:i-1} \Vert y_i} \rangle$
    \EndFor
    \State $B_{i+1} \gets$ Max $B_{t+i}$
\EndFor
\State \textbf{Output:} $\arg\max B_T$
\end{algorithmic}
\label{alg:beam}
\end{algorithm}

Let the token set at scale \( i \) be denoted as \( \mathcal{V}(i) \). We identify a subset of \( n \) tokens, denoted as \( \mathcal{V}_{\text{choose}}(i) \), based on the following criterion:

\begin{equation}
\mathcal{V}_{\text{choose}}(i) = \underset{v \in \mathcal{V}(i)}{\text{top-}n} \left( {P_{\text{max}}(v)}+\alpha  ({P_{\text{max}}(v) - P_{\text{2nd}}(v))} \right)
\end{equation}

Here, \(P_{\max}(v)\) and \(P_{\mathrm{2nd}}(v)\) denote the highest and second-highest predicted probabilities for token \(v\), respectively. The coefficient \(\alpha\) is a tunable hyperparameter. This selection prioritizes tokens with low confidence and high uncertainty. For each of the \( n \) selected tokens, we consider both optimal and suboptimal candidate values, resulting in \( 2^n \) possible token configurations for scale \( i \). 
These candidate sets are propagated through the autoregressive decoder to predict tokens at scale \( i{+}1 \). Scoring function is then used to rank all candidates, defined as:

\begin{equation}
\text{Score} = s + \sum \log(P_{\text{max}}(\hat{y}))
\end{equation}

where \( s \) is the accumulated score from previous steps, and \( \hat{y} \) represents the predicted tokens at scale i+1. The highest-scoring configuration is retained for subsequent inference.

This method effectively combines the scale-wise progressive generation and scoring mechanism of \mymethod{} model, while reducing computational complexity through selective search. It ensures high-quality generated sequences and alleviates the issue of error accumulation to some extent.

\section{Experiments}

\subsection{Experimental Setup}

\textbf{Datasets.} We utilize BraTS2021 (1,251 MRI volumes) \cite{Bakas2017} and LIDC-IDRI (1,308 CT scans) \cite{armato2011lung}. Data is split into training, validation, and testing sets (7:1:2). Slices are cropped to $256 \times 256$, with a $4\times$ upsampling factor. Low-resolution (LR) images are synthesized using Gaussian blur ($\sigma=1$).

\textbf{Baselines.} We benchmark \mymethod{} against representative state-of-the-art (SOTA) models, encompassing both multi-contrast (MCSR: Disc-Diff \cite{mao2023disc}, DiffMSR \cite{li2024rethinking}) and single-contrast (SCSR: SwinIR \cite{liang2021swinir}, LDM \cite{rombach2022high}, MMHCA \cite{georgescu2023multimodal}, VARSR \cite{qu2025visual}) paradigms.
On BraTS2021, we evaluate both SCSR and MCSR (T1-guided T2 reconstruction) setups; LIDC-IDRI is restricted to SCSR.

\textbf{Implementation Details.} The framework is implemented in PyTorch and trained on four NVIDIA RTX 4090 GPUs for 100 epochs using the Adam optimizer. Training proceeds in two stages: (1) initializing the VAE and discriminator (batch size 24; learning rates $2\times10^{-4}$ and $1\times10^{-4}$); (2) training the full autoregressive model (batch size 32; learning rate $1\times10^{-4}$). More details can be found in the appendix.

\subsubsection{Metric.} We assess reconstruction quality using both reference-based and no-reference metrics. PSNR and SSIM are standard full-reference metrics evaluating pixel-level fidelity and structural similarity, respectively. LPIPS and DISTS \cite{ding2020image} measure deep feature differences using pretrained networks, offering perceptual insights. For no-reference evaluation, MUSIQ \cite{ke2021musiq} and MANIQA \cite{yang2022maniqa} capture multi-scale and attention-based perceptual quality.

\begin{table*}[t]
\centering
% \small
% \scriptsize
\caption{\textbf{Quantitative results on Brats and Lung dataset.}}
\begin{tabular}{c|l|c|c|cccccc}
\toprule
Dataset & Method & Data Type & Model Type & PSNR$\uparrow$ & SSIM$\uparrow$ & LPIPS$\downarrow$ & DISTS$\downarrow$ & MUSIQ$\uparrow$ & MANIQA$\uparrow$    \\
\midrule
\multirow{8}{*}{\textit{Brats}} 
& Disc-Diff\cite{mao2023disc} & MC & Diffusion & 33.57        & 0.9241       &  \BLUE{0.1002} & 0.1497& 39.34& 
0.2511  \\
& DiffMSR\cite{li2024rethinking}   & MC & Regression & \RED{34.49}  & \BLUE{0.9318}& 0.1991       & 0.2145       & 27.35       & 0.2439\\
& Bicubic   & SC & Interpolation & 29.11        & 0.8382       & 0.4470       & 0.3185       & 18.01       & 0.1438\\
& SwinIR\cite{liang2021swinir}    & SC & Regression & 33.33        & 0.9161       & 0.1302       & 0.1674       & 34.75       & 0.2756\\
& MMHCA\cite{georgescu2023multimodal}     & SC & Regression & 33.48        & 0.9230       & 0.1316       & 0.1693       & 35.84       & 0.2888\\
& LDM\cite{rombach2022high}       & SC & Diffusion & 32.43        & 0.9156       & 0.1544       & 0.1762      & 34.14        & 0.2689 \\
& VARSR \cite{qu2025visual} & SC & Autoregressive & 33.65 & 0.9270 & 0.1127 & \BLUE{0.1185}  & \BLUE{44.38} & \BLUE{0.3249} \\
& MIAR(Ours)      & SC & Autoregressive &\BLUE{34.22} & \RED{0.9367} & \RED{0.0556}  & \RED{0.0773} & \RED{47.87} & \RED{0.3431}  \\
\midrule
\multirow{5}{*}{\textit{Lung}} 
& Bicubic   & SC & Interpolation & 33.07          & 0.8426        & 0.4481        & 0.2965       & 21.49        & 0.1457         \\
& MMHCA\cite{georgescu2023multimodal}     & SC & Regression &\BLUE{35.77}   & \BLUE{0.9072} & 0.1470 & 0.1602 & 33.93 & 0.2487 \\
& SwinIR\cite{liang2021swinir}     & SC &  Regression &35.13          & 0.8926        & 0.1526        & 0.1736       & 32.55        & 0.2423       \\
& LDM\cite{rombach2022high}        & SC & Diffusion & 34.80          & 0.8713        & 0.1997        & 0.1885       & 26.10        & 0.2204     \\
& VARSR \cite{qu2025visual} & SC & Autoregressive & 35.70 & 0.9070 & \BLUE{0.1394} & \BLUE{0.1094} & \BLUE{36.80} &  \BLUE{0.2504} \\
& MIAR(Ours)      & SC  & Autoregressive & \RED{36.39} & \RED{0.9171} & \RED{0.0779} & \RED{0.0840} & \RED{40.87} & \RED{0.2674}  \\
\bottomrule
\end{tabular}

\vspace{-3px}
\label{tab:Quantitative_all}
\end{table*}

\subsection{Qualitative Analysis}

Fig.~\ref{fig:contrast} compares super-resolution results on BraTS and LIDC-IDRI datasets, with magnified insets (yellow/green boxes) highlighting local details. While DisC-Diff produces perceptually plausible images, it suffers from low structural fidelity and anatomical hallucinations. DiffMSR preserves fine-grained features but lacks perceptual sharpness. SwinIR yields over-smoothed outputs, losing subtle textures and edges, while LDM produces blurred reconstructions with limited structural clarity. Although the AR-based VARSR exhibits high perceptual quality, its diffusion-based refinement stage introduces slight blurriness. In contrast, our controllable autoregressive method achieves superior perceptual realism and high  structural fidelity, yielding results nearly indistinguishable from the ground truth. More visual results are available in the appendix.

\subsection{Quantitative Analysis}

Table~\ref{tab:Quantitative_all} evaluates \mymethod{} on the BraTS (MRI) and LIDC-IDRI (CT) datasets, where it consistently surpasses all single-contrast (SCSR) baselines. Although multi-contrast (MCSR) methods such as DiffMSR exhibit a higher PSNR on BraTS by leveraging auxiliary inputs, \mymethod{} achieves superior perceptual realism, dominating all competitors in both reference-based and no-reference perceptual metrics.

Non-autoregressive paradigms often succumb to the perception-fidelity trade-off, characterized by blurry regression-based outputs (SwinIR, MMHCA) or anatomically inconsistent diffusion hallucinations (DisC-Diff, LDM). \mymethod{} reconciles the long-standing fidelity-perception trade-off. Specifically, it achieves a 44.51\% reduction in LPIPS and a 21.68\% gain in MUSIQ while simultaneously elevating PSNR and SSIM. Compared to the AR-based VARSR \cite{qu2025visual}, \mymethod{} establishes new SOTA benchmarks for all reference-based and no-reference metrics across both datasets. These results underscore \mymethod{}’s capacity to reconcile structural fidelity with vivid perceptual realism.

As shown in Tab.~\ref{tab:beam_search}, AR-based methods outperform diffusion-based approaches in generation speed, though they remain slower than regression-based methods.

\begin{table}[htbp]
\centering
\small
\caption{\textbf{Ablation on Hierarchical Beam Search}}
\begin{tabular}{l|ccc}
\toprule
Type & PSNR$\uparrow$ & SSIM$\uparrow$ & Time$\downarrow$ \\
\midrule
Disc-Diff\cite{mao2023disc}         & 33.57 & 0.9241 & 1455ms \\
LDM\cite{rombach2022high}           & 32.43 & 0.9156 & 832ms  \\
VARSR \cite{qu2025visual}             & 33.65 & 0.9270 & 390ms  \\
topK                                & 33.49 & 0.9287 & 257ms  \\
topP                                & 33.52 & 0.9292 & 266ms  \\
Beam Search n=1                     & 34.22 & 0.9367 & 412ms  \\
Beam Search n=2                     & 34.25 & 0.9369 & 705ms  \\
Beam Search n=3                     & 34.27 & 0.9369 & 1372ms \\
\bottomrule
\end{tabular}
\label{tab:beam_search}
\end{table}

\subsection{Ablation Study}

% # 前两个考虑画到一张表上。后面一个可以画一个小表，考虑采用环绕的方式？
\subsubsection{ROPE}

\begin{figure}[htb] 
\centering 
\includegraphics[width=0.9\linewidth]{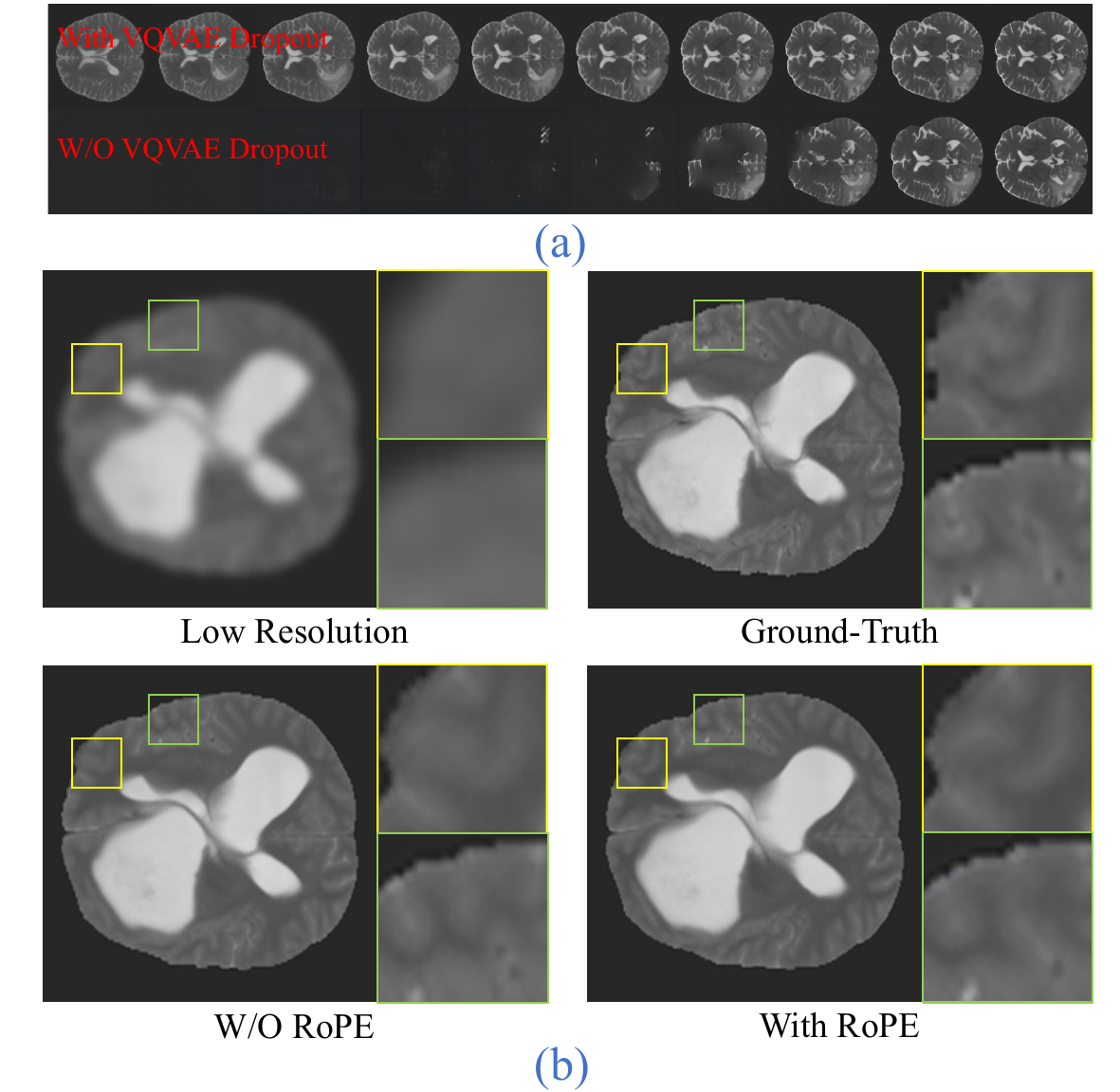} 
\caption{
\textbf{Visual analyse on ablation studies.}
(a) scale-wise reconstructions from MSVQ-VAE with and without scale-aware dropout. The dropout improves intermediate consistency and structural coherence.
(b) Visual comparison of reconstructed slices with and without RoPE, where RoPE improves anatomical fidelity and preserves fine-grained textures in medical images. 
}
\vspace{-10px}
\label{fig:twoimage}
\end{figure}

As shown in Tab.~\ref{tab:other_aba}, the incorporation of RoPE significantly enhances the performance on reference-based metrics. This improvement indicates that RoPE enables the model to better preserve fine-grained structural details and semantic fidelity in the reconstructed HR images. Moreover, visual comparisons in Fig.~\ref{fig:twoimage}b further validate that RoPE facilitates superior retention of anatomical features, which is critical for clinical applications requiring high-precision image interpretation.

\begin{table}[htbp]
\centering
\small
\caption{\textbf{Ablation on ROPE and Scale-Aware Dropout}}
\begin{tabular}{l|cc}
\toprule
Type & PSNR$\uparrow$ & SSIM$\uparrow$  \\
\midrule
w/o Scale-Aware Dropout  & 32.78 & 0.9024 \\
w/o Rope  & 33.45 & 0.9210 \\
MIAR(ours) & 34.22 & 0.9367 \\
\bottomrule
\end{tabular}
\label{tab:other_aba}
\end{table}

\subsubsection{Scale-Aware Dropout}

As shown in Tab.~\ref{tab:other_aba}, applying dropout in the MSVQ-VAE bottleneck notably improves reference-based metrics such as PSNR and SSIM, indicating that regularized latent representations facilitate higher-fidelity reconstructions. From Fig.~\ref{fig:twoimage}a, we observe that models trained with scale-aware dropout tend to incrementally enrich structural and textural details across successive scales. In contrast, when dropout is disabled, the model defers most of the information reconstruction to the final few scales, leading to an imbalanced distribution of semantic content throughout the multi-scale hierarchy.

\begin{figure}[htb] 
\centering 
\includegraphics[width=0.9\linewidth]{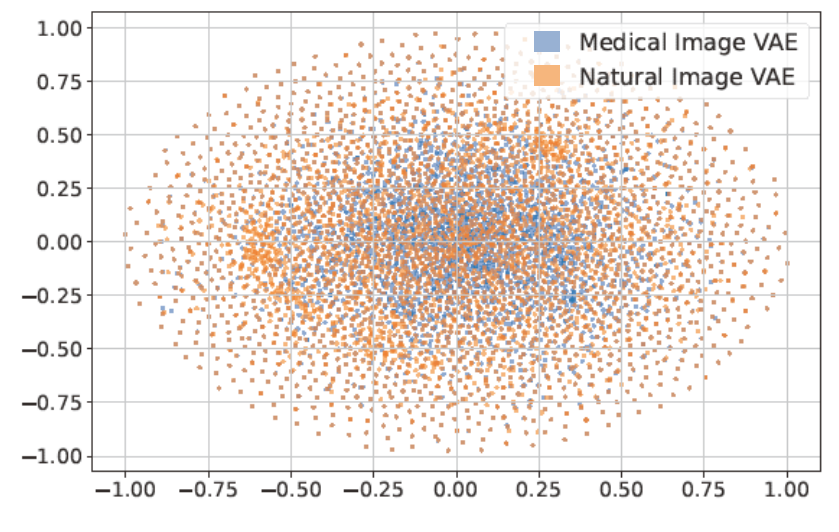} 
\caption{
\textbf{t-SNE visualization.}
Visual comparison of reconstructed slices with and without RoPE, where RoPE improves anatomical fidelity and preserves fine-grained textures in medical images. 
}
\vspace{-10px}
\label{fig:TSNE}
\end{figure}

\subsubsection{Hierarchical Beam Search}
\label{subsec:beam_search}

As shown in Tab.~\ref{tab:Quantitative_all} and Tab.~\ref{tab:beam_search}, incorporating beam search with a beam width of $n=1$ (equivalent to selecting two candidates per scale)   yields substantial improvements in reconstruction quality. Although further increasing $n$ can still lead to marginal gains, the performance benefit quickly diminishes relative to the associated computational overhead during inference.

To further analyze hierarchical beam search, we visualize codebook distributions using T-SNE (Fig.~\ref{fig:TSNE}). Compared to natural images with dispersed embeddings, medical images exhibit a more compact latent space, indicating stronger structural consistency but higher susceptibility to sampling errors in autoregressive models like VAR. Hierarchical beam search alleviates this by preserving multiple high-probability paths, reducing error accumulation during decoding.
\section{Conclusion and Disscusion}
\label{sec:conclusion}
We propose \mymethod{}, a multi-scale autoregressive framework for medical image super-resolution. By enhancing the VAR backbone with Cross-Attention Fusion and Rotary Positional Encoding, and introducing a hierarchical beam search decoding strategy, our model improves semantic consistency, spatial fidelity, and inference stability. \mymethod{} achieves superior performance on perceptual quality metrics and competitive results on pixel-wise fidelity measures, while significantly outperforming diffusion-based models in inference speed. This highlights its effectiveness and practicality for high-resolution medical image super-resolution.

\section{Acknowlegemetns}

This work was supported in part by the National Natural Science Foundation of China (82272576 and 62572032), the Shandong Natural Science Foundation (ZR2025MS1106), Disciplines Breakthrough Plan of the Ministry of Education of China  (JYB2025XDXM610), the Guangxi Science and Technology Major Program (GuiKeAA24206017),  the Beijing Nova Program (20250484786), the Beijing Natural Science Foundation (4252018), and the Fundamental Research Funds for the Central Universities.
{
\bibliographystyle{IEEEbib}
\bibliography{icme2026references}

@article{feng2022multimodal,
  title={Multimodal transformer for accelerated MR imaging},
  author={Feng, Chun-Mei and Yan, Yunlu and Chen, Geng and Xu, Yong and Hu, Ying and Shao, Ling and Fu, Huazhu},
  journal={TMI},
  volume={42},
  number={10},
  pages={2804--2816},
  year={2022},
  publisher={IEEE}
}

@inproceedings{khaledyan2020low,
  title={Low-cost implementation of bilinear and bicubic image interpolation for real-time image super-resolution},
  author={Khaledyan, Donya and Amirany, Abdolah and Jafari, Kian and Moaiyeri, Mohammad Hossein and Khuzani, Abolfazl Zargari and Mashhadi, Najmeh},
  booktitle={2020 IEEE Global Humanitarian Technology Conference (GHTC)},
  pages={1--5},
  year={2020},
  organization={IEEE}
}

@inproceedings{zhang2021mr,
  title={MR image super-resolution with squeeze and excitation reasoning attention network},
  author={Zhang, Yulun and Li, Kai and Li, Kunpeng and Fu, Yun},
  booktitle={CVPR},
  pages={13425--13434},
  year={2021}
}

@article{li2021high,
  title={High-resolution pelvic MRI reconstruction using a generative adversarial network with attention and cyclic loss},
  author={Li, Guangyuan and Lv, Jun and Tong, Xiangrong and Wang, Chengyan and Yang, Guang},
  journal={IEEE Access},
  volume={9},
  pages={105951--105964},
  year={2021},
  publisher={IEEE}
}

@inproceedings{li2024rethinking,
  title={Rethinking diffusion model for multi-contrast mri super-resolution},
  author={Li, Guangyuan and Rao, Chen and Mo, Juncheng and Zhang, Zhanjie and Xing, Wei and Zhao, Lei},
  booktitle={CVPR},
  pages={11365--11374},
  year={2024}
}

@inproceedings{mao2023disc,
  title={Disc-diff: Disentangled conditional diffusion model for multi-contrast mri super-resolution},
  author={Mao, Ye and Jiang, Lan and Chen, Xi and Li, Chao},
  booktitle={MICCAI},
  pages={387--397},
  year={2023},
  organization={Springer}
}

@inproceedings{ji2024deform,
  title={Deform-mamba network for mri super-resolution},
  author={Ji, Zexin and Zou, Beiji and Kui, Xiaoyan and Vera, Pierre and Ruan, Su},
  booktitle={MICCAI},
  pages={242--252},
  year={2024},
  organization={Springer}
}

@article{tian2024visual,
  title={Visual autoregressive modeling: Scalable image generation via next-scale prediction},
  author={Tian, Keyu and Jiang, Yi and Yuan, Zehuan and Peng, Bingyue and Wang, Liwei},
  journal={NeurIPS},
  volume={37},
  pages={84839--84865},
  year={2024}
}

@inproceedings{liang2021swinir,
  title={Swinir: Image restoration using swin transformer},
  author={Liang, Jingyun and Cao, Jiezhang and Sun, Guolei and Zhang, Kai and Van Gool, Luc and Timofte, Radu},
  booktitle={ICCV},
  pages={1833--1844},
  year={2021}
}

@inproceedings{rombach2022high,
  title={High-resolution image synthesis with latent diffusion models},
  author={Rombach, Robin and Blattmann, Andreas and Lorenz, Dominik and Esser, Patrick and Ommer, Bj{\"o}rn},
  booktitle={CVPR},
  pages={10684--10695},
  year={2022}
}

@article{sun2024autoregressive,
  title={Autoregressive model beats diffusion: Llama for scalable image generation},
  author={Sun, Peize and Jiang, Yi and Chen, Shoufa and Zhang, Shilong and Peng, Bingyue and Luo, Ping and Yuan, Zehuan},
  journal={arXiv preprint arXiv:2406.06525},
  year={2024}
}

@inproceedings{qu2025visual,
  title={Visual Autoregressive Modeling for Image Super-Resolution},
  author={Qu, Yunpeng and Yuan, Kun and Hao, Jinhua and Zhao, Kai and Xie, Qizhi and Sun, Ming and Zhou, Chao},
  booktitle={ICML},
  pages={50926--50948},
  year={2025},
  organization={PMLR}
}

@inproceedings{georgescu2023multimodal,
  title={Multimodal multi-head convolutional attention with various kernel sizes for medical image super-resolution},
  author={Georgescu, Mariana-Iuliana and Ionescu, Radu Tudor and Miron, Andreea-Iuliana and Savencu, Olivian and Ristea, Nicolae-C{\u{a}}t{\u{a}}lin and Verga, Nicolae and Khan, Fahad Shahbaz},
  booktitle={WACV},
  pages={2195--2205},
  year={2023}
}

@article{armato2011lung,
  title={The lung image database consortium (LIDC) and image database resource initiative (IDRI): a completed reference database of lung nodules on CT scans},
  author={Armato III, Samuel G and McLennan, Geoffrey and Bidaut, Luc and McNitt-Gray, Michael F and Meyer, Charles R and Reeves, Anthony P and Zhao, Binsheng and Aberle, Denise R and Henschke, Claudia I and Hoffman, Eric A and others},
  journal={Medical physics},
  volume={38},
  number={2},
  pages={915--931},
  year={2011},
  publisher={Wiley Online Library}
}

@article{Bakas2017,
  author = {Bakas, Spyridon and Akbari, Hamed and Sotiras, Aristeidis and Bilello, Michel and Rozycki, Martin and Kirby, Justin S. and Freymann, John B. and Farahani, Keyvan and Davatzikos, Christos},
  year = {2017},
  date = {2017/09/05},
  title = {Advancing The Cancer Genome Atlas glioma MRI collections with expert segmentation labels and radiomic features},
  journal = {Scientific Data},
  pages = {170117},
  volume = {4},
  number = {1},
  issn = {2052-4463},
  url = {https://doi.org/10.1038/sdata.2017.117},
  doi = {10.1038/sdata.2017.117},
}

@article{ding2020image,
  title={Image quality assessment: Unifying structure and texture similarity},
  author={Ding, Keyan and Ma, Kede and Wang, Shiqi and Simoncelli, Eero P},
  journal={PAMI},
  volume={44},
  number={5},
  pages={2567--2581},
  year={2020},
  publisher={IEEE}
}

@inproceedings{ke2021musiq,
  title={Musiq: Multi-scale image quality transformer},
  author={Ke, Junjie and Wang, Qifei and Wang, Yilin and Milanfar, Peyman and Yang, Feng},
  booktitle={ICCV},
  pages={5148--5157},
  year={2021}
}

@inproceedings{yang2022maniqa,
  title={Maniqa: Multi-dimension attention network for no-reference image quality assessment},
  author={Yang, Sidi and Wu, Tianhe and Shi, Shuwei and Lao, Shanshan and Gong, Yuan and Cao, Mingdeng and Wang, Jiahao and Yang, Yujiu},
  booktitle={CVPR},
  pages={1191--1200},
  year={2022}
}

@inproceedings{li2024imagefolder,
  title={ImageFolder: Autoregressive Image Generation with Folded Tokens},
  author={Li, Xiang and Qiu, Kai and Chen, Hao and Kuen, Jason and Gu, Jiuxiang and Raj, Bhiksha and Lin, Zhe},
  booktitle={ICLR}
}
}

% \input{section/supply.tex}
% \input{section/5_conclusion_disscusion}
% {
% % \bibliographystyle{abbrv}
% \bibliographystyle{IEEEbib}
% \bibliography{icme2026references}
% }

\end{document}